\documentclass{article}

\usepackage{times}
\usepackage{soul}
\usepackage[utf8]{inputenc}
\usepackage{graphicx}
\usepackage{amsmath}
\usepackage[table]{xcolor}
\usepackage{floatrow}

\title{\LARGE \bf
Identifying Driver Behaviors using\\ Trajectory Features for Vehicle Navigation
}

\author{Ernest Cheung$^{1}$, Aniket Bera$^{1}$, Emily Kubin$^{2}$, \\
Kurt Gray$^{2}$, and Dinesh Manocha$^{1}$
\\
\texttt{http://gamma.cs.unc.edu/TDBM }
\thanks{$^{1}$Authors from the Department of Computer Science, University of North Carolina at Chapel Hill, USA}
\thanks{$^{2}$Authors from the Department of Psychology and Neuroscience, University of North Carolina at Chapel Hill, USA
}%
}

\begin{document}

\maketitle
\thispagestyle{empty}
\pagestyle{empty}

\begin{abstract}
We present a novel approach to automatically identify driver behaviors from vehicle trajectories and use them for safe navigation of autonomous vehicles. We propose a novel set of features that can be easily extracted from car trajectories. We derive a data-driven mapping between these features and six driver behaviors using an elaborate web-based user study.  We also compute a summarized score indicating a level of awareness that is needed while driving next to other vehicles.  We also incorporate our algorithm into a vehicle navigation simulation system and demonstrate its benefits in terms of safer real-time navigation, while driving next to aggressive or dangerous drivers.

\end{abstract}

\section{Introduction}


Identifying dangerous drivers is crucial in developing safe autonomous driving algorithms and advanced driving assistant systems.  The problem has been extensively studied in transportation and urban planning research \cite{meiring2015review}.  However, prior work usually correlates driver' behaviors with their backgrounds (e.g., driver age, response to questionnaires, etc.).  On the other hand, to develop autonomous vehicle systems, we need to understand the behavior of surrounding drivers using only the sensor data.  As with to a human driver, an autonomous navigation algorithm that can predict other vehicle's driving behavior can navigate safely and efficiently avoid getting near dangerous drivers.

Prior work in transportation research \cite{feng2012selected, meiring2015review} often characterizes drivers using their levels of aggressiveness and carefulness.  Several works in modeling pedestrian trajectories \cite{guy2011simulating} and navigation \cite{bera2017aggressive} algorithms have applied psychological theory to capture human behavior. 
Current autonomous driving systems uses a range of different algorithms to process sensor data.  Object detection and semantic understating methods are applied to obtain trajectory data \cite{Geiger2012CVPR}. Some work \cite{bojarski2016end} uses end-to-end approaches to make navigation decisions from the sensor inputs (e.g. camera images, LIDAR data, etc.).  

\textbf{Main Results:}
We present a novel approach to automatically identifying driver behaviors from vehicle trajectories. We perform an extensive user study to learn the relationship and establish a mathematical mapping between extracted vehicular trajectories and the underlying driving behaviors: Trajectory to Driver Behavior Mapping (TDBM).  TDBM enables a navigation algorithm to automatically classify the driving behavior of other vehicles.  We also demonstrate simulated scenarios where navigating with our improved navigation scheme is safer.

Our approach takes into account different trajectory features.   We use five different features, which can be easily extracted from vehicle trajectories and used to classify driving behaviors. We show that selecting a subset of these features is more favorable than selecting the currently used ones to produce a strong regression model that maps to driving behaviors.



As compared to prior algorithms, our algorithm offers the following benefits:

\textbf{1. Driving Behavior Computation: } We present a data-driven algorithm to compute TDBM. We conducted a comprehensive user survey to establish a mapping between five features and six different driving behaviors.  We further conduct factor analysis on the six behaviors, which are derived from two commonly studied behaviors: aggressiveness and carefulness.  The results show that there exists a latent variable that can summarize these driving behaviors and that can be used to measure the level of awareness that one should have when driving next to a vehicle.  In the same study, we examine how much attention a human would pay to such a vehicle when it is driving in different relative locations.

\textbf{2. Improved Realtime Navigation}:  We compute the features and identify the driving behaviors using TDBM. We enhance an existing Autonomous Driving Algorithm \cite{best2017autonovi} to navigate according to the neighboring drivers' behavior.  Our navigation algorithm identifies potentially dangerous drivers in realtime and chooses a path that avoids potentially dangerous drivers.    

\begin{figure*}[!ht]
\centering
\fbox{\includegraphics[width=0.98\textwidth]{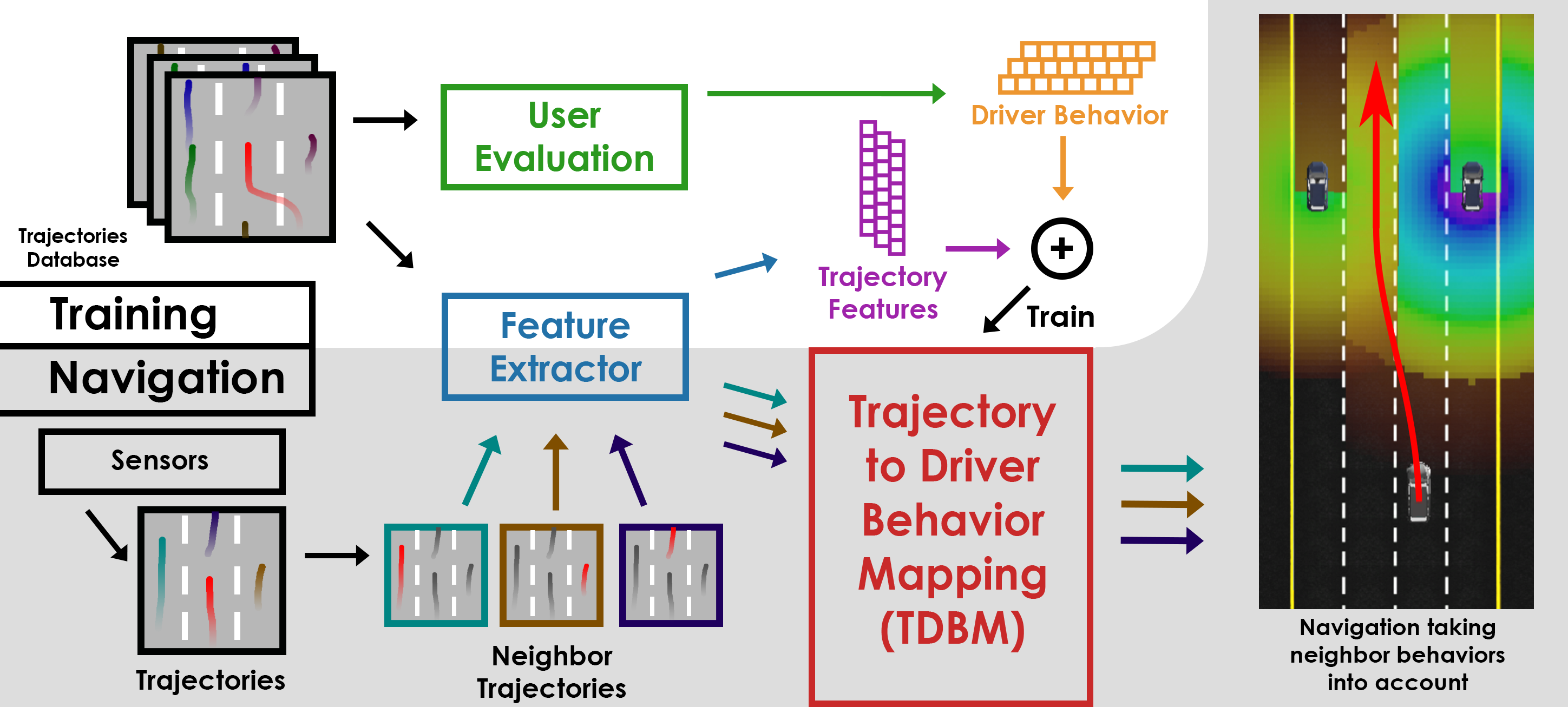}}

\caption{Overview of our Algorithm:  During the training of TDBM, we extract features from the trajectory database and conduct a user evaluation to find the mapping between them.  During the navigation stage, we compute a set of trajectory and extract the features, then compute the driving behavior using TDBM.  Finally, we plan for real-time navigation, taking into account these driver behaviors.   
}
\label{fig:outline}
\end{figure*}

An overview of our approach is shown in Figure \ref{fig:outline}.  The rest of the paper is organized as follows. We give a brief overview of prior work in Section \ref{sec:rw}. We introduce the new trajectory features that are used to identify the driver behaviors in Section \ref{sec:method}. We present our data-driven mapping algorithm (TDBM) in Section \ref{sec:ana} and use it for autonomous car navigation in Section \ref{sec:nav}.

\section{Related Works} \label{sec:rw}

\subsection{Studies on Driving Behaviors} \label{sec:rwds}
There has been a wide range of work studying drivers' behaviors in Social Psychology and Transportation.  Feng et al. \cite{feng2012selected} proposed five driver characteristics (age, gender, year of driving experience, personality via blood test, and education level) and four environmental factors (weather, traffic situation, quality of road infrastructure, and other cars' behavior), and mapped them to 3 levels of aggressiveness (driving safely, verbally abusing other drivers, and taking action against other drivers).  Aljaafreh et al. \cite{aljaafreh2012driving} categorized driving behaviors into 4 classes: Below normal, Normal, Aggressive, and Very aggressive, in accordance to accelerometer data.  Social Psychology studies \cite{krahe2002predicting,beck2014distress} have examined the aggressiveness according to the background of the driver, including age, gender, violation records, power of cars, occupation, etc.  Mouloua et al. \cite{brill2009predictive} designed a questionnaire on subjects' previous aggressive driving behavior, and concluded that these drivers also repeated those behaviors under a simulated environment.  Meiring et al. \cite{meiring2015review} used several statistical reports to conclude that distracted behaviors and drunk behaviors are also serious threats to road safety.  
Many of the driver features used by these prior methods cannot be easily computed in new, unknown environments using current sensors.   Our work uses trajectory data which can be extracted from sensor data in most autonomous driving systems.

\subsection{Trajectories Features} \label{sec:rwf}
Murphey et al. \cite{murphey2009driver} conducted an analysis on the aggressiveness of drivers and found that longitudinal (changing lanes) jerk is more related to aggressiveness than progressive (along the lane) jerk (i.e. rate of change in acceleration).  Mohamad et al. \cite{mohamad2011abnormal} detected abnormal driving styles using speed, acceleration, and steering wheel movement, which indicate direction of vehicles. 
Qi et al. \cite{qi2015leveraging} studied driving styles with respect to speed and acceleration. Shi et al. \cite{shi2015evaluating} pointed out that deceleration is not very indicative of aggressiveness of drivers, but measurements of throttle opening, which is associated with acceleration, is more helpful in identifying aggressive drivers.  Wang et al. \cite{wang2017driving} classified drivers into two categories, aggressive and normal, using speed and throttle opening captured by a simulator.  

Sadigh et al. \cite{sadigh2014data} proposed a data-driven model based on Convex Markov Chains to predict whether a driver is paying attention while driving.  There are considerable works on development of in-car smart systems to alert users when they are found driving distracted by indicating nearby vehicles \cite{iOnRoad}, departures from lane markers and drivers' appearances and road conditions \cite{you2013carsafe}, and trajectories computed by camera, IMU and GPS \cite{bergasa2014drivesafe}.

Instead of directly analyzing real-world data, many methods model driving behaviors as input parameters to generate driving simulations. Treiber et al. \cite{treiber2000congested} proposed a lane following model, that controls the speed of the car using desired velocity, minimum spacing, desired time headway, acceleration, and maximum breaking deceleration.  Kesting et al. \cite{kesting2007general} proposed a lane changing model, that makes lane changing decisions based on the speed advantage gained and the speed disadvantage imposed on the other vehicles, using a franticness and a politeness factor.  Choudhury et al. \cite{choudhury2005modeling} proposed a complex lane changing model, composed of desired speed, desired time gap, jam distance, maximum acceleration, desired deceleration, coolness factor, minimum acceptable gap, etc. 

 We combine a set of selected features proposed by previous works in terms of behavior mapping and simulation with two new trajectory features, lane following metric and relative speed metric.  Then, we use variable selection to select a subset of features that can produce a good regression model. 
 
\subsection{Autonomous Car Navigation}
There is substantial work on autonomous vehicle navigation~\cite{katrakazas2015real,saifuzzaman2014incorporating,kolski2006autonomous,hoffmann2007autonomous}.  Ziegler et al. \cite{ziegler2014making} presented a navigation approach that is capable of navigating through the historic Bertha Benz route in Germany. Numerous navigation approaches \cite{buehler2009darpa, geiger2012team, kianfar2012design, englund2016grand} have been proposed in the DAPRA Urban Grand Challenge and the Grand Cooperative Driving Challenge.   Recent work proposed by Best et al. \cite{best2017autonovi}, AutonoVi, presented an improved navigation algorithm that takes into account dynamic lane changes, steering and acceleration planning, and various other factors. Our approach is complimentary to these methods and can be combined with them. 


\subsection{Adaptation to Human Drivers' Behavior}
Sadigh et al. \cite{sadigh2016planning} observed that an autonomous car's action could also affect neighboring human drivers' behavior, and studied how humans will react when the autonomous car performs certain actions \cite{sadigh2016information}.  Huang et al. \cite{huang2017enabling} presented techniques for making autonomous car actions easily understandable to humans drivers. They also proposed an active learning approach \cite{dorsa2017active} to model human driving behavior by showing examples of how a human driver will pick their preference out of a given set of trajectories.  While this stream of work went further to take into account how humans would react to an autonomous car's action, it also emphasized the importance of a robot navigating according to other drivers' behavior.  

\section{Methodology} \label{sec:method}
In this section, we present the two novel trajectory features that are used to identify driver behaviors. We also compare their performance with other features and give an overview of driver behavior metrics used in our navigation algorithm.  






\subsection{Features} \label{sec:fea}
The goal of our work is to extract a set of trajectory features that can be mapped properly to driving behaviors.  We assume that the trajectories have been extracted from the sensor data.  Many of the previous works deal with different driver characteristics: driver background, accelerometer use, throttle opening, etc., which may not be available for an autonomous vehicle in new and uncertain environments.   Moreover, in the simulation models described in Section \ref{sec:rwf}, a lot of features cannot be measured from trajectories with insufficient lane-changing samples: comfortable breaking deceleration, desired time headway, etc.  Therefore, we derive some variants of features that can be easily extracted from the trajectories and summarize them in Table \ref{tab:fea}.  These features are further shortlisted with the results from a user study described in the next section.

\begin{table}[h]
\centering
\scalebox{0.8}{
\begin{tabular}{c !{\vrule width -1pt}l !{\vrule width -1pt}l}

\textbf{Symbol} & \textbf{Notation}  & \textbf{Description}                                               \\
\hline
$f_0$ & $v_{front}$   & Average relative speed to the car in front                     \\
$f_1$ & $v_{back}$   & Average relative speed to the car in the back                  \\
$f_2$ & $v_{left}$   & Average relative speed to cars in the left lane             \\
$f_3$ & $v_{right}$   & Average relative speed to cars in the right lane            \\
\rowcolor[HTML]{759AE5}$f_4$ & $v_{nei}$ & Relative speed to neighbors \\[-0.5pt]
\rowcolor[HTML]{759AE5}$f_5$ & $v_{avg}$   & Average velocity        \\[-0.5pt]
\rowcolor[HTML]{00B271}$f_6$ & $s_{front}$   & Distance with front car  \\[-0.5pt]
\rowcolor[HTML]{00B271}$f_7$ & $j_{l}$ & Longitudinal jerk \\
$f_8$ & $j_{p}$ & Progressive jerk \\
\rowcolor[HTML]{759AE5}$f_9$ & $s_{center}$ & Lane following metric

\end{tabular}
}
\caption{We considered ten candidate features ${f_0, .., f_9}$ for selection.  Features highlighted in green are selected for mapping to behavior-metrics only, and those in blue are selected for mapping to both behavior-metrics and attention metrics. } 
\label{tab:fea}
\end{table}

\subsubsection{Acceleration}
As pointed out in several prior works \cite{murphey2009driver, mohamad2011abnormal, shi2015evaluating, wang2017driving}, acceleration is often correlated with driver aggressiveness.  While previous studies \cite{murphey2009driver} concluded that longitudinal jerk can reflect aggressiveness better than progressive jerk, our goal is to use features that also correlate with all the driving styles, instead of just aggressiveness.  Therefore, we include both longitudinal jerk $j_l$ and progressive jerk $j_p$ in our computations.  

\subsubsection{Lane following}
Previous work \cite{bergasa2014drivesafe} proposed a metric measureing the extent of lane following that depends on the mean and standard deviation of lane drifting and lane weaving.  We propose a feature that also depends on lane drifting, but distinguishes between drivers who keep drifting left and right within a lane and those who are driving straight but not along the center of the lane.   Moreover, we compensate for the extent of lane drifting while performing lane changing to avoid capturing normal lane changing behaviors into this metric. 

Given $y_{l}$, which is the center longitudinal position of the lane that the targeted car is in, and $y(t)$, which is the longitudinal position of the car at time $t$,  we detect a lane changing event when the car has departed from one lane to the another and remained in the new lane for at least $k$ seconds. 

With a set of changing lane events happened at time $t_i$, $C = \{t_1, t_2, ..., t_n\}$, the lane drift metric $s_{C}(t)$ is measured as below:

\begin{equation}
s_{C}(t) = 
\begin{cases}
  0, & \text{if} \, \exists t \in C \, \text{s.t.} \, t \in [t-k,t+k], \\
  y(t) - y_{l}, & \text{otherwise}.
\end{cases}
\end{equation}

\begin{figure}[!ht]
\centering

\includegraphics[width=0.98\textwidth]{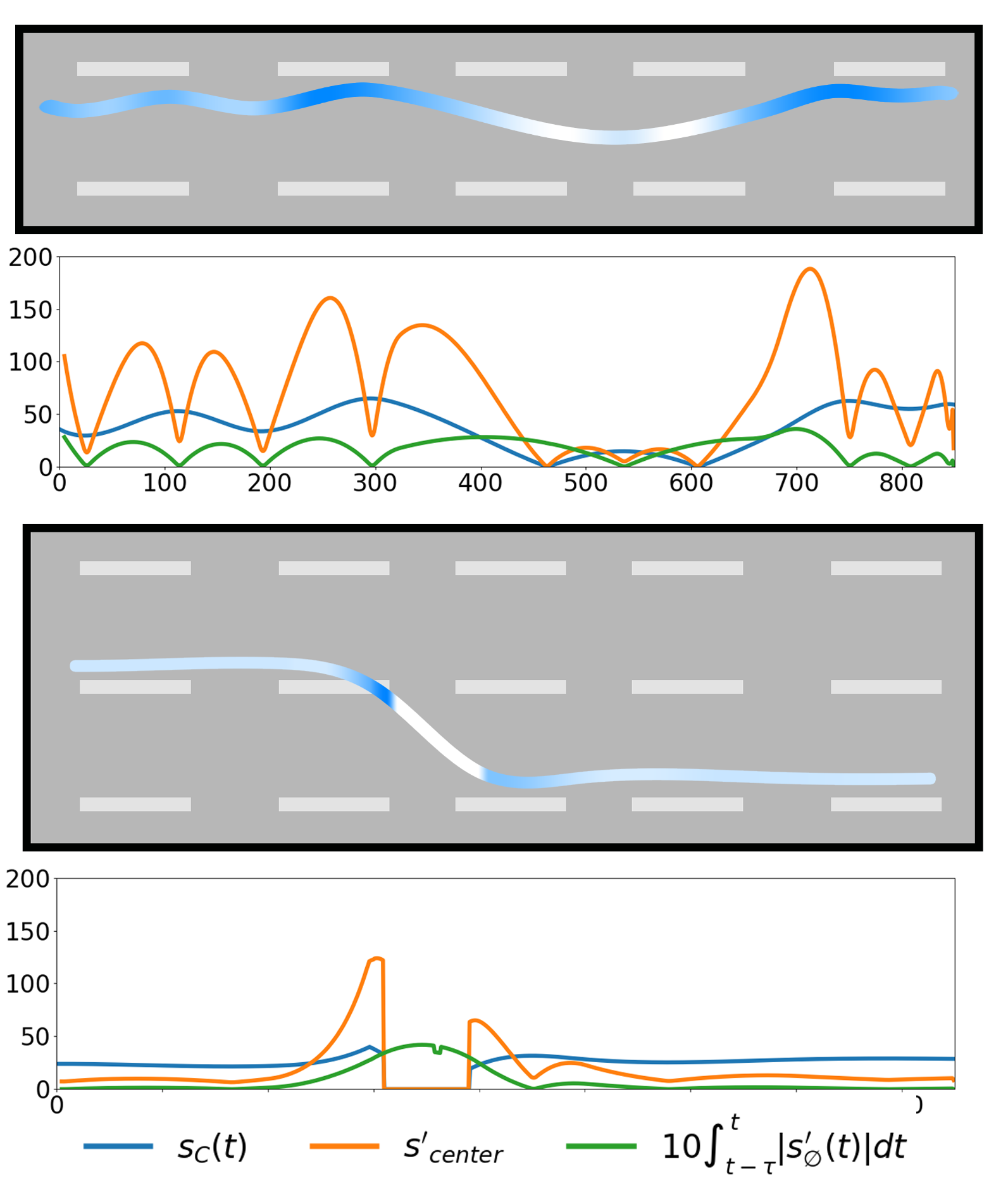}

\caption{  Illustration of the lane drift metric ($|s_c(t)|$), and the lane following metric ($s_{center}$).  The lane following metric for the trajectories above is the sum of the area under the plot of $s'_{center}$.  This two example shows that our lane following metric ($s_{center}$) captures the `drifting behavior' in the top example, but not the `driving straight off the center' and `lane changing' shown in the bottom example.} 
\label{fig:drift}

\end{figure}

We use a term that measures the previous $\tau$ seconds of rate of change in drifting to differentiate lane drifts from those drivers who are driving straight but off the center of the lane.  Our overall lane following metric is illustrated in Figure \ref{fig:drift} and defined as:
\begin{equation}
s_{center} = \int |s_{C}(t)| \bigg[\mu + \int_{t-\tau}^{t} |s_{\emptyset}'(t)| dt \bigg] \, dt,
\label{eq:cen}
\end{equation}
where $\mu$ is a parameter that distinguish drivers who are driving off the center of the lane and those who are along.

\subsubsection{Relative Speed} \label{sec:rsn}
Relative speed has been used to evaluate the aggressiveness of drivers \cite{qi2015leveraging}.  However, directly measuring the relative speed using $v_{front}$, $v_{back}$, $v_{left}$ and $v_{right}$ has many issues.  First, such a feature sometimes does not exist as there may be no car next to the target car.  Second, these features might not be directly related to the driving behavior of the car.  While driving substantially faster than other cars would be perceived as aggression, driving slower might not necessarily imply that the driver is non-aggressive.  Third, computing such an average velocity requires knowledge about the trajectories and range of speeds of the neighboring vehicles. Given these considerations, we design the following metric to capture the relationship between the driving behavior and the relative speed with respect to neighboring cars:

\begin{equation}
v_{nei} = \int \sum_{n \in N} \max (0, \frac{ v(t) - v_{n}(t)}{dist(x(t) , x_n(t))}) dt,
\label{eq:vnei}
\end{equation}
where $N$ denotes the set containing all neighboring cars within a large range (e.g., a one-mile radius). $x(t)$, $v(t)$, $x_{n}(t)$, $v_{n}(t)$ denote the position and the speed of the targeting car, and the position and the speed of the neighbor n, respectively.



\subsection{Driving Behavior Metrics} \label{sec:dsm}

As discussed in Section \ref{sec:rwds}, aggressiveness \cite{feng2012selected,aljaafreh2012driving,HARRIS20141} and carefulness \cite{meiring2015review,sadigh2014data,lan2009smartldws} are two metrics that have been used to identify road safety threats. Typically, social psychologists add related items into studies to leverage robustness and the observed effects.  Therefore, we would like to evaluate four more driving behaviors: Reckless, Threatening, Cautious, and Timid. They are listed in Table \ref{tab:per}. 






\subsection{Attention Metrics} \label{sec:am}
Observing different maneuvers of other drivers on the road can result in paying more attention to those drivers. However, the relative position of such drivers (with respect to the targeted vehicle) would affect the level of attention that one is paying to them.  For instance, one would pay more attention to a vehicle in the front making frequent stops, as opposed to a following vehicle.  We would like to understand how much attention a driver will pay to the targeted car when the user assumes that he or she is driving in different relative positions than the target.  We study four different relative positions: preceding, following, adjacent to and far away from the targeted vehicle, also listed in Table \ref{tab:per}.  These positions affect the level of attention one would pay when driving in that relative position.

\begin{table}[h]
\centering
\scalebox{0.8}{
\begin{tabular}{cl|cl}
\textbf{Symbol} &  \textbf{Description} & \textbf{Symbol} & \textbf{Level of Attention when} \\
\hline
$b_0$ & Aggressive & $b_6$ &  following the target  \\
$b_1$ & Reckless & $b_7$ &   preceding the target \\
$b_2$ & Threatening & $b_8$ & driving next to the target  \\
$b_3$ & Careful & $b_9$ &  far from the target\\
$b_4$ & Cautious \\
$b_5$ & Timid      

\end{tabular}
}
\caption{Six Driving Behavior metrics ($b_0$, $b_1$, ...,$b_5$) and Four Attention metrics ($b_6$, $b_7$, $b_8$, $b_9$) used in TDBM }

\label{tab:per}
\end{table}



\section{Data-Driven Mapping}\label{sec:ana}
We designed a user study, involving 100 participants to identify driver behaviors from videos rendered from the Interstate 80 Freeway Dataset \cite{i80}.  The video dataset consists of 45 minutes of vehicle trajectories, captured in a 1650 feet section on I-80 in California, US.  The videos were first annotated automatically using a proprietary code developed in the NGSIM program, and then manually checked and corrected.  The raw videos provided in the dataset are low in quality and divided into seven different segments with different camera angles.  Therefore, we have rendered the videos using a game engine, Unreal Engine, to provide a stable and consistent view for the users in the survey. The virtual cameras have a fixed transform to the targeted car, which is highlighted in red, and will follow it throughout the video.  

\begin{figure}[!ht]
\centering
\includegraphics[width=\textwidth]{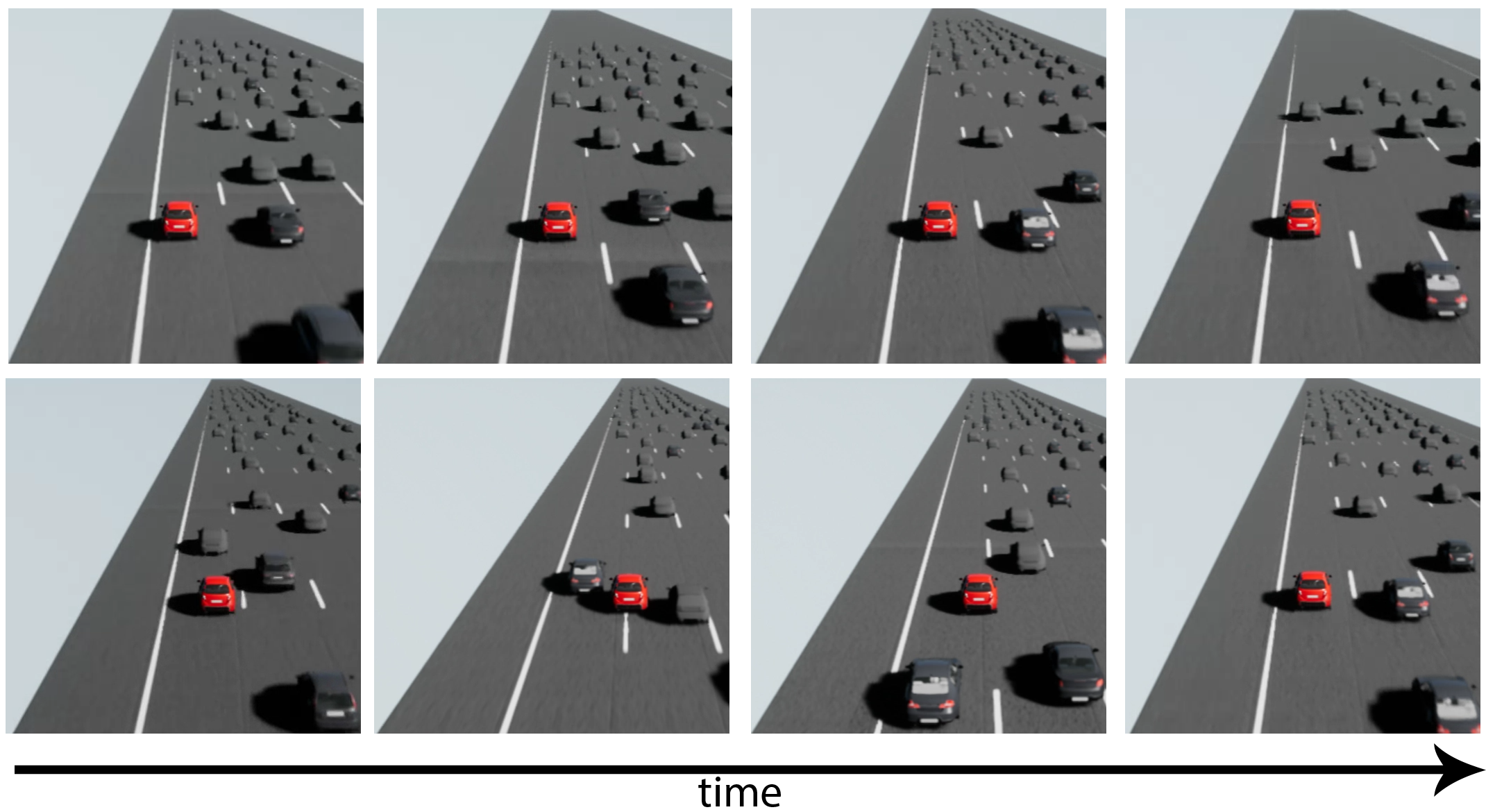}

\caption{  Two example videos used in the user study.  Participants are asked to rate the six driving behavior metrics and four attention metrics of the target car colored in red.  }
\label{fig:ussnap}
\end{figure}

Figure \ref{fig:ussnap} shows snapshots of the videos used in the user study.  The participants were asked to rate the six behaviors we described in Section \ref{sec:dsm} on a 7-point scale: \{Strongly disagree, Disagree, Somewhat disagree, Neither agree or disagree, Somewhat agree, Agree, Strongly agree\}.  This was followed by another question on how much attention they would be paying if they were in different positions relative to the targeted car, as described in Section \ref{sec:am}, on a 5-point scale, where -2 indicates not at all, 0 indicates a moderate amount and 2 indicates a lot.

\subsection{Data Pre-Processing} \label{sec:dataug}
We perform data augmentation to make sure that the dataset has a sufficiently wide spectrum of driving behaviors corresponding to lane changes, fast moving cars, passing cars, etc.  In addition, the features described in Table \ref{tab:fea} are measured using different units.  To improve numerical stability during the regression analysis, we scale the data linearly using the 5th and the 95th percentile samples.

\subsection{Feature Selection} \label{sec:fs}
In Section \ref{sec:fea} and Table \ref{tab:fea}, we cover a wide range of features used in previous studies that can be extracted from trajectories, along with two new metrics that attempt to summarize some of these features to avoid strong correlation between independent variables during regression analysis.  In this section, we apply feature selection techniques to find out which features are most relevant to the driving behaviors.  

We perform least absolute shrinkage and selection operator (Lasso) analysis on six driving behaviors $b_0$, $b_1$, ..., $b_5$ and four attention metrics, $b_6$, $b_7$, $b_8$, $b_9$, from the user responses.  The objective function for Lasso analysis conducted on $b_i$ is:

\begin{equation}
    \min_{\beta_{i}',\beta_i} \Big[ \frac{1}{N} \sum_{j=1}^{N}(b_i - \beta{i}' - f_j^T \beta_{i,j}) \Big] \, \text{, subject to} \, \sum_{j=1}^{F} |\beta_{i,j}| \leq \alpha_i,
    \label{eq:lasso}
\end{equation}

\noindent where $N$ is the number of survey responses and $F$ is the number of features.  

\begin{figure}[!ht]
\centering
\includegraphics[width=\textwidth]{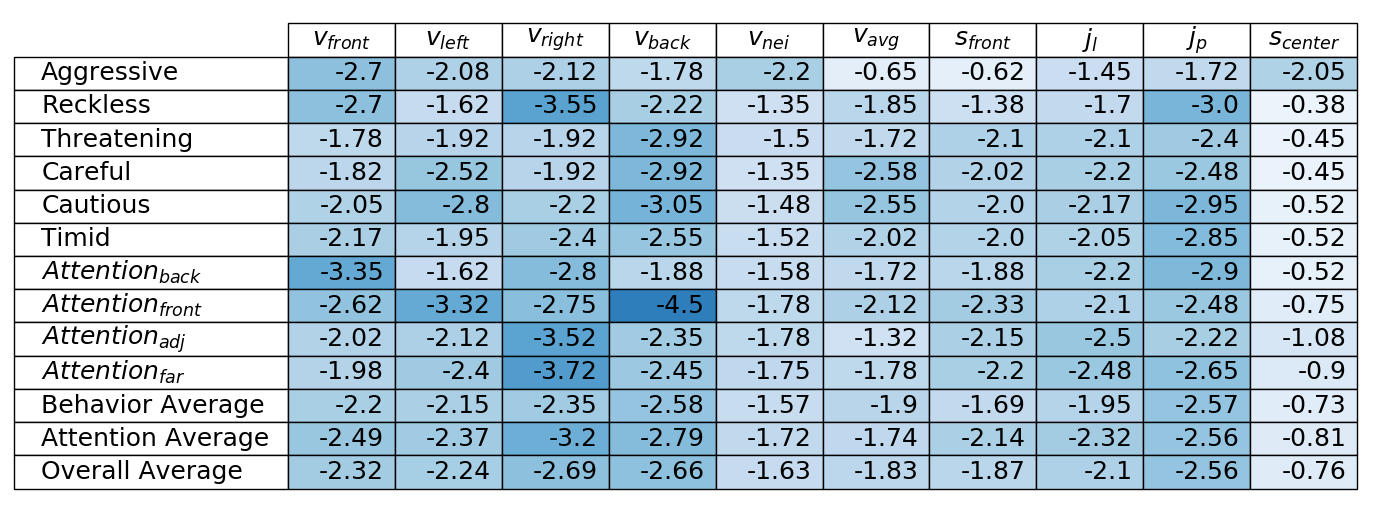}

\caption{ The converging value (in the power of 10) of $\beta_{i,j}$ which maps a feature $f_{j}$ to a behavior/attention metric $b_{i}$ while performing Lasso analysis.  A larger converging value indicates a higher likelihood that the feature is favourable in regression analysis, and therefore we select that value for TDBM. }
\label{fig:fs}
\end{figure}

Lasso analysis performs regularization and feature selection by eliminating weak subsets of features.  The parameter $\alpha_i$ determines the level of regularization that Lasso analysis imposes on the features.  As we increase $\alpha_{i}$, features $f_j$ will be eliminated in a different order.  Unlike regular regression analysis on a single dependent variable, our goal is to select two sets of features: one that can produce a strong regression model for all six driving behavior metrics, and one for all four attention metrics.  We sample different values of $\alpha_{i}$ for all responses $b_i$, and record the values of $\alpha_{i}$ at which the component $\beta_{i,j}$ (which mapping feature $f_j$ to response $b_i$) converges to 0. The results are shown in Figure \ref{fig:fs}, where converging values of $\beta_{i,j}$ are presented in the power of 10.

The directly computed relative speeds of the cars surrounding the targeted car are least favorable for selection for both regressions for behavior-metrics and attention-metrics.  However, our relative speed metric proposed to capture the correlation between surrounding cars and the targeted car, $v_{nei}$ (Equation \ref{eq:vnei}), is more favorable in terms of being selected. Moreover, our lane following metric, $s_{center}$ (Equation \ref{eq:cen}), tends to be the last one eliminated as a feature in the variable selection stage.

Our goal is to find two $\alpha_{behavior}$ and $\alpha_{attention}$ that shortlist a subset of features for behavior-metric and attention-metric respectively.  Note that $\alpha_{behavior}=\alpha_{i} ,\, \forall i \in [0,5]$, and $\alpha_{attention}=\alpha_{i} ,\, \forall i \in [6,9]$ for $\alpha_{i}$ defined in Equation \ref{eq:lasso}.  In terms of behavior, we can either pick $\{ s_{center}, v_{nei}, s_{front} \}$ or $\{ s_{center}, v_{nei}, s_{front}, v_{avg}, j_{l} \}$.  Given that the mapping component between $v_{avg}$ and $j_{l}$ has high converging values, they can produce a stronger regression model for aggression, and that aggressiveness is one of the common behaviors as studied in prior literature discussed in Section \ref{sec:rwds}.  We therefore select the latter set of features for behavior mapping.  For mapping features with attention regions metrics, we select $\{ s_{center}, v_{nei},  v_{avg} \}$.

\subsection{Feature-Behavior Mapping}

Using $\{ s_{center}, v_{nei}, s_{front}, v_{avg}, j_{l} \}$ and $\{ s_{center}, v_{nei},  v_{avg} \}$ as the features, we perform linear regression to obtain the mapping between these selected features and the drivers' behavior.  We normalize the data as described in Section \ref{sec:dataug} to increase the numerical stability of the regression process.  The results we obtained are below.  For $B_{behavior} = [b_{0}, b_{1}, ..., b_{5}]^T$, we obtain

\begin{equation}
\scalebox{0.7}{\mbox{\ensuremath{\displaystyle{
   B_{behavior} = 
    \begin{pmatrix} 
    1.63  &  4.04 & -0.46 & -0.82 &  0.88 & -2.58 \\
    1.58  &  3.08 & -0.45 &  0.02 & -0.10 & -1.67 \\
    1.35  &  4.08 & -0.58 & -0.43 & -0.28 & -1.99 \\
    -1.51 & -3.17 &  1.06 &  0.51 & -0.51 & 1.39 \\
    -2.47 & -2.60 &  1.43 &  0.98 & -0.82 & 1.27 \\
    -3.59 & -2.19 &  1.75 &  1.73 & -0.30 & 0.61 
    \end{pmatrix}
    \begin{pmatrix} 
    s_{center} \\
    v_{nei} \\
    s_{front} \\ 
    v_{avg} \\
    j_{l} \\
    1
    \end{pmatrix}
    }}}}
    \label{eq:beh}
\end{equation}

Moreover, for $B_{attention} = [b_{6}, b_{7}, b_{8}, b_{9}]^T$, 

\begin{equation}
\scalebox{0.7}{\mbox{\ensuremath{\displaystyle{
   B_{attention} = 
   \begin{pmatrix}
   B_{back} \\
   B_{front} \\
   B_{adj} \\
   B_{far} 
    \end{pmatrix} =
    \begin{pmatrix} 
     0.54 &  1.60 &  0.11 & -0.8 \\
    -0.73 & 1.66 & 0.63 & -0.07 \\
    -0.14 & 1.73 & 0.25 & 0.15 \\
    0.25 & 1.47 & 0.17 & -1.43 
    \end{pmatrix}
    \begin{pmatrix} 
    s_{center} \\
    v_{nei} \\
    v_{avg} \\
    1
    \end{pmatrix}
    }}}}
    \label{eq:att}
\end{equation}

We further apply leave-one-out cross-validation to the set of samples $S$: enumerate through all samples $s_{i} \in S$ and leave $s_{i}$ as a validation sample, and use the remaining samples $S - s_{i}$ to produce regression models $M_{i,j}$ for each behavior $b_{i,j}$.  Using $M_{i,j}$, we predict the behaviors $b_{i,j}$ of $s_{i}$.  The mean prediction errors of $b_{i,j}$ using $M_{i,j}$ are listed in the table below.  The mean prediction error in the cross-validation is less than 1 in a 7-point scale for all behaviors and attention metrics predicted, showing that our mappings described in Equation \ref{eq:beh} and \ref{eq:att} are not over-fitted.

\begin{table}[h]
\centering
\scalebox{0.8}{
\begin{tabular}{c|c|c|c|c|c|c|c|c|c}
$b_0$ & $b_1$ & $b_2$ & $b_3$ & $b_4$ & $b_5$ & $b_6$ & $b_7$ & $b_8$ & $b_9$ \\ \hline
0.75 & 0.94 & 0.78 & 0.7 & 0.6 & 0.89 & 0.2 & 0.49 & 0.38 & 0.23 
\label{tab:err}
\end{tabular}
}
\caption{ Mean error in a 7-point scale when applying cross validation of linear regression to map feature to behavior and attention metrics showing our mapping is not over-fitted.  }
\end{table}

\subsection{Factor Analysis}
Previous studies on mapping walking behavior adjectives with features used to simulate crowds \cite{guy2011simulating},
have applied factor analysis to find smaller numbers of primary factors that can represent the personalities or behaviors. 
We can apply Principal Component Analysis (PCA) to the survey response.  The percentages of variance of the principal components are 73.42\%, 11.97\%, 7.78\%, 2.96\%, 2.30\% and 1.58\%.  The results indicate that the Principal Component 1, which has variance of 73.43\%, can model most of the driving behaviors. 

\begin{figure}[!ht]
\centering
\begin{tabular}{cc}
\includegraphics[width=0.48\textwidth]{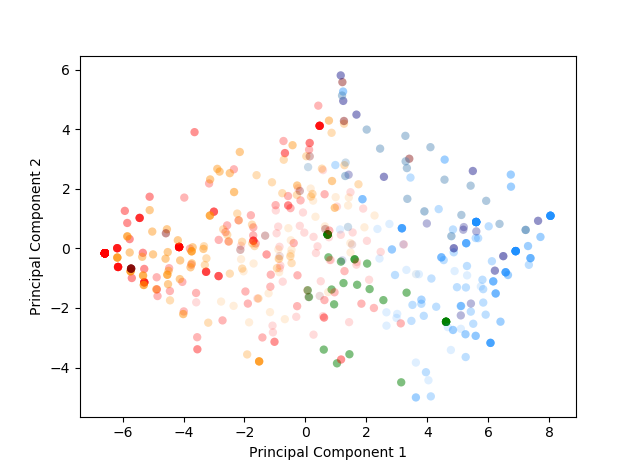} &
\includegraphics[width=0.48\textwidth]{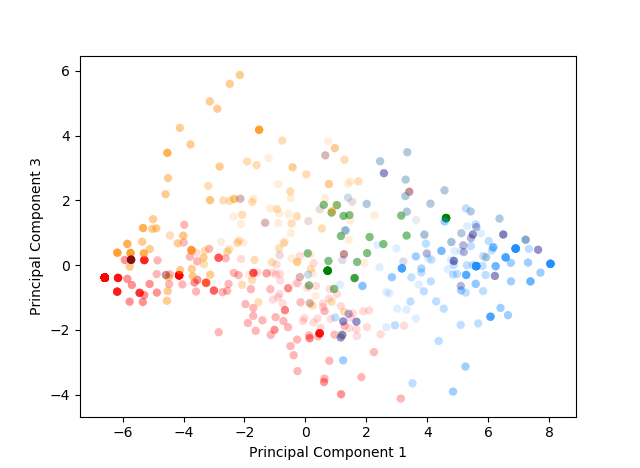} \\
\multicolumn{2}{c}{\includegraphics[width=\textwidth]{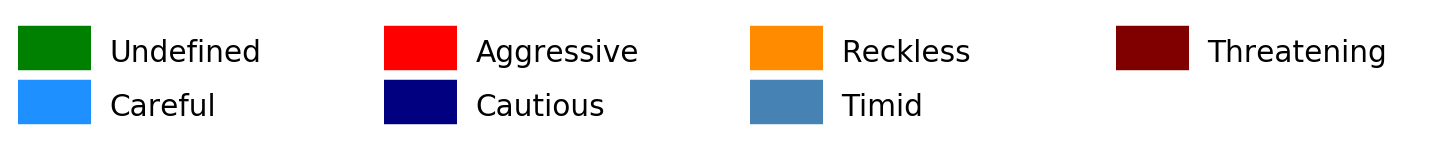}}
\end{tabular}
\caption{ Principal Component Analysis results for \{Principal Components 1(PC1), PC2\} (left) and \{PC1, PC3\} (right).  The color of the data point indicates the highest rated driving behavior adjective as shown in the legends, and the alpha value indicates the rating of this behavior (solid for `Strongly agree', and half-transparent for `Somewhat agree').  If a user did not agree to any of the behaviors or indicated multiple contradicting behaviors, the data point is marked as undefined in green.    }
\label{fig:fa}
\end{figure}

We represent each entry of the user study response with the highest rated behavior and transform these entries into the space of the Principal Components as shown in Figure \ref{fig:fa}.  If the user did not fully agree to any behavior for a video (i.e. responses to all questions are below `Somewhat agree'), we consider that there to be no representative behavior from this entry (i.e. undefined).  Also, if a response indicates more than one behavior as the strongest, then we label those behaviors as undefined if those adjectives contradict each other (i.e. one from negative adjectives \{Aggressive, Reckless, Threatening\} and one from positive adjectives \{Careful, Cautious, Timid\}).  As observed in Figure \ref{fig:fa}, the distribution of the data on Principal Component 1, the three negative behavior adjectives we used in the user study, represented in warmer colors, are distributed on the negative side, while the three positive behavior adjectives are distributed on the positive side.  Furthermore, the entries that suggest the users' responses were `Strongly agree', represented by solid color plots in Figure \ref{fig:fa}, have significantly higher magnitudes in terms of value along Principal Component 1.  However, for Principal Components 2 and 3, such a relationship is not observed.

\begin{figure}[!ht]
\centering
\begin{tabular}{cc}
\includegraphics[width=0.48\textwidth]{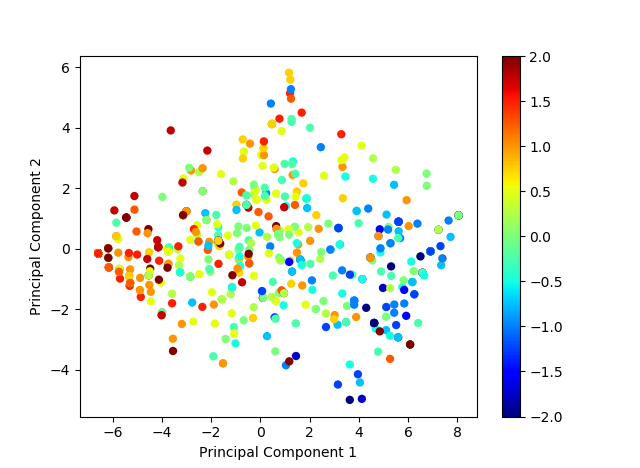} &
\includegraphics[width=0.48\textwidth]{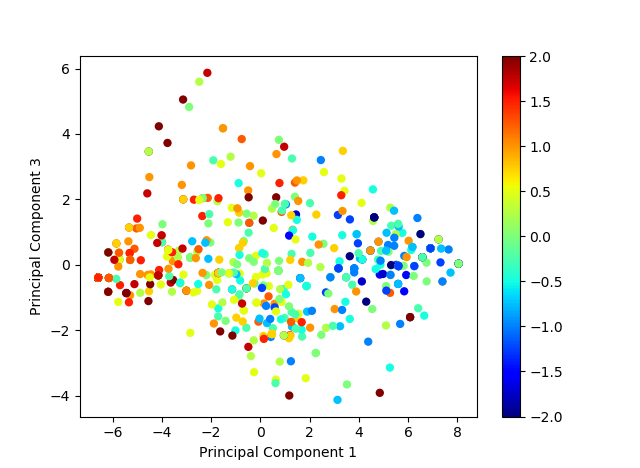} \\
\end{tabular}
\caption{ Principal Component Analysis results for \{PC1, PC2\} (left) and \{PC1, PC3\} (right).  The color of the data point indicates the average amount of awareness the user rated on a 5 point scale (-2 for not paying any attention at all, and 2 for paying a lot of attention).   }
\label{fig:fa2}
\end{figure}

Our studies show that there could be one latent variable that is negatively correlated with aggressiveness and positively correlated with carefulness.  We further verify these results by analyzing the correlation of the  Principal Components with the amount of awareness that the users indicated they would pay to the targeted car.  We take the average of the level of attention, $\frac{b_6+b_7+b_8+b_9}{4}$, recorded for each response and plot these averages as the color on the PCA results in Figure \ref{fig:fa2}.  Similar results have been observed from this user evaluation, where the drivers worth more attention have a lower value of Principal Component 1, and those who worth less attention tend to have a higher value.  Moreover, there is no clear evidences pointing to correlation between the level of awareness the user rated and Principal Component 2 or 3.

Therefore, we consider the Principal Component 1 as a safety score reflecting the amount of attention awareness that a driver or an autonomous navigation system should take into account.  TDBM is therefore computed as below:

\begin{equation}
\scalebox{0.7}{\mbox{\ensuremath{\displaystyle{
   S_{TDBM} = 
    \begin{pmatrix} 
     -4.78 & -7.89 & 2.24 &  1.69 & -0.83 & 4.69 \\
    \end{pmatrix}
   \begin{pmatrix} 
    s_{center} \\
    v_{nei} \\
    s_{front} \\ 
    v_{avg} \\
    j_{l} \\
    1
    \end{pmatrix}
    }}}}
\label{eq:sums}
\end{equation}

\section{Navigation} \label{sec:nav}

In this section, we highlight the benefits of identifying driver behaviors and how these ensure safe navigation. We extend an autonomous car navigation algorithm, AutonoVi \cite{best2017autonovi}, and show improvements in its performance by using our driver behavior identification algorithm and TDBM. AutonoVi is based on a data-driven vehicle dynamics model and optimization-based maneuver planning, which generates a set of favorable trajectories from among a set of possible candidates, and performs selection among this set of trajectories using optimization.  It can handle dynamic lane-changes and different traffic conditions.  

The approach used in AutonoVi is summarized below:  The algorithm takes a graph of roads from a GIS database, and applies A* algorithm to compute the shortest global route plan.  The route plan consists of a sequence of actions that is composed of \{Drive Straight, Turn Left, Turn Right, Merge Left, and Merge Right\}.  The plan is translated to a static guiding path that consists of a set of way-points, that exhibits $C^1$ continuity, and that takes Traffic Rules into account (e.g., making a stop at an intersection).  AutonoVi then samples the steering angle and velocity in a desirable range of values to compute a set of candidate trajectories, and eliminates the trajectories that lead to possible collisions based on Control Obstacles \cite{bareiss2015generalized}.  

Among the set of collision-free trajectories, AutonoVi selects the best trajectory by optimizing a heuristic that penalizes trajectories that lead to: i) deviation from global route; ii) sharp turns, braking and acceleration; iii) unnecessary lane changes; and iv) getting too close to other vehicles and objects (even without a collision).  

To avoid getting too close to other neighboring entities, AutonoVi proposed a proximity cost function to differentiate entities only by its class.  That is, it considers all vehicles as the same and applies the same penalization factor, $F_{vehicle}$, to them.  Further, it applies a higher factor : $F_{ped}$ and $F_{cyc}$ to pedestrians and cyclist respectively.  The original proximity cost used in AutonoVi is:

\begin{equation}
c_{prox} = \sum_{n=1}^{N} F_{vehicle} \, e^{-d(n)}
\end{equation}

This cost function has two issues: i) it cannot distinguish dangerous drivers to avoid driving too close to them, and ii) it diminishes too rapidly due to its use of an exponential function. We propose a novel proximity cost that can solve these problems:

\begin{equation}
    c'_{prox} = \sum_{n=1}^{N} c(n) 
\end{equation}
    
\begin{equation}
\scalebox{0.85}{\mbox{\ensuremath{\displaystyle{
c(n) = \begin{cases}
0 & \text{if} \, d \in [d_{t2}, \inf) , \\[3pt]
S_{TDBM} B_{far} \frac{d_{t2} - d(n)}{d_{t2}} & \text{if} \, d \in (d_{t},  d_{t2}],  \\[3pt]
S_{TDBM} \big[\frac{(d_{t} - d(n))(B_{r}-B_{far})}{d_{t}} +B_{far}\big] & \text{if} \, d \in (0,d_{t}] .
    
\end{cases}
}}}}
\end{equation}

\noindent where $d(n)$ is the distance between the car navigating with our approach and the neighbor $n$, $d_{t}$ is a threshold distance beyond which neighbors are considered as far away, and $d_{t2}$ is a threshold distance beyond which neighbors would no longer have impact on our navigation.  $S_{TDBM}$ is derived from Equation \ref{eq:sums}, $B_{far}$ and $B_{r}$ are the attention metrics computed using the features extracted from the features using the mapping in equation \ref{eq:att}, for $r$ = \{back, front, adj\} if the neighboring car is following, preceding, and next to the navigating car, respectively.

\begin{figure}[!ht]
\centering
\begin{tabular}{cc}
\includegraphics[width=0.97\textwidth]{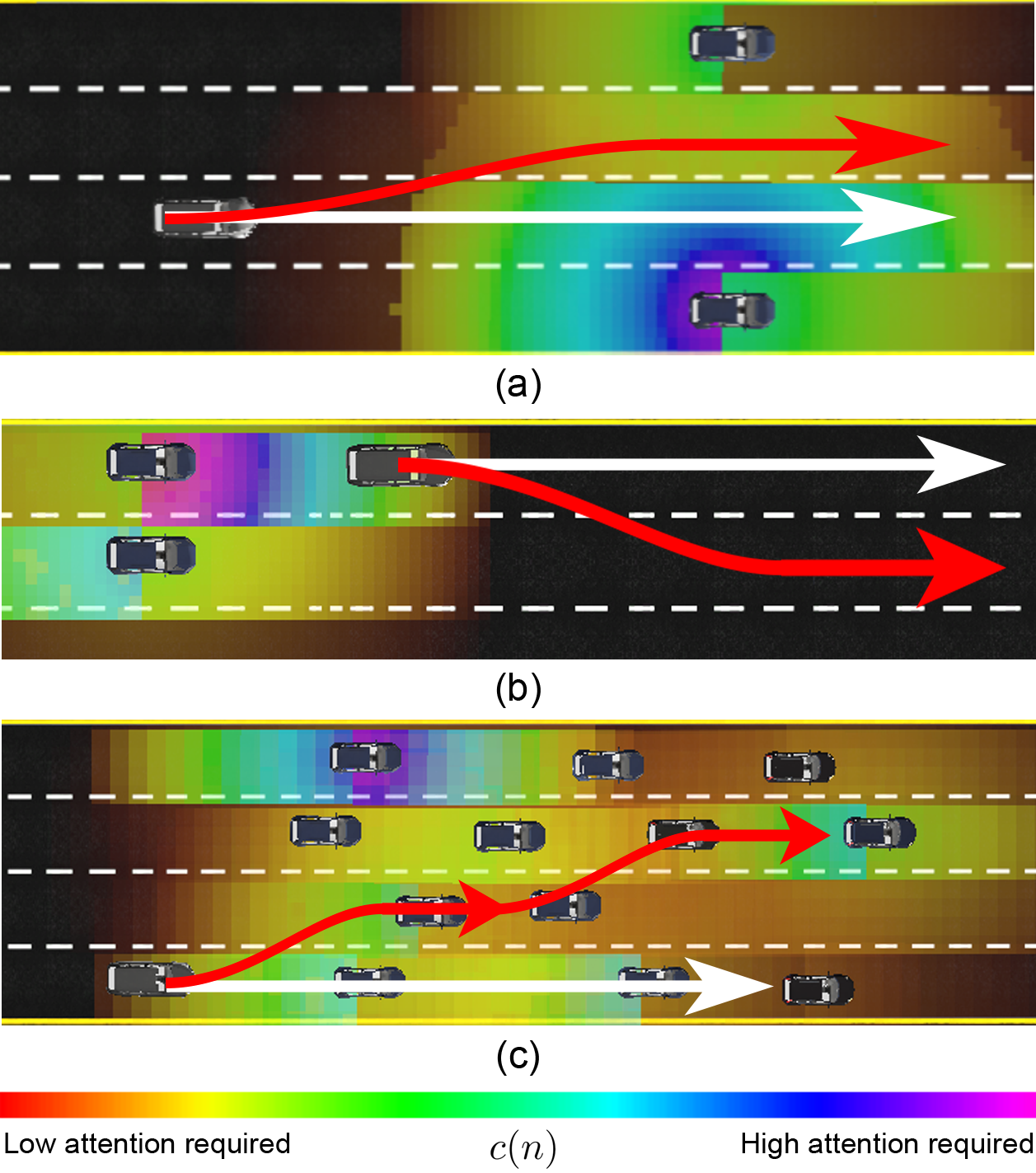}
\end{tabular}
\caption{Examples of our approach making better navigation decision than AutonoVi.  The red route is the one selected by our approach while white is the one selected by AutonoVi.  The cost map $c(n)$ is also shown for each neighbor car $n$ indicating the amount of attention needed.  In (a), our algorithm chooses to switch lane and keep a distance from the car require more attention.  In (b), a car requiring high level of attention tailgates the car running our approach, and we switch to a slower lane to give way.  In (c), a heavy traffic ahead causing all four lanes move at a similarly low speed, and our algorithm chooses to the follow the car with the lowest attention required.  }
\label{fig:nav}
\end{figure}

Using this new cost function, we can avoid drivers that are potentially riskier, and select a safer navigation path.  Examples of scenarios are illustrated in Figure \ref{fig:nav} and the attached video.  

\section {Conclusion and future works }
We present a novel data-driven approach to enable safer real-time navigation by identifying human drivers who are potentially hazardous.  Our studies and findings are based on a data-driven mapping computation (TDBM).  We conclude that although humans use different adjectives when describing driving behavior, there is an underlying latent variable, $S_{TDBM}$  (Equation \ref{eq:sums}), that reflects the level of attention humans pay to other vehicles' driving behavior.  Moreover, we can estimate this variable by a set of novel trajectory features and other existing features.  

Current trajectory data tends to be limited due to human labeling or the fact that extra efforts may be needed to extract such annotated data from raw images. With advancement in object detection and other work in computer vision, one can expect more trajectory data would be made available to the autonomous driving research community. Given more variety of data (e.g., in urban environments or different cultures), we would like to apply our approach to analyzing and developing different navigation strategies that adapt to these new situations and local driving styles.

\section{Acknowledgement}
This research is supported in part by ARO grant W911NF16-1-0085, and Intel.

\bibliographystyle{IEEEtran}
\bibliography{iros}

\begin{thebibliography}{10}
\providecommand{\url}[1]{#1}
\csname url@samestyle\endcsname
\providecommand{\newblock}{\relax}
\providecommand{\bibinfo}[2]{#2}
\providecommand{\BIBentrySTDinterwordspacing}{\spaceskip=0pt\relax}
\providecommand{\BIBentryALTinterwordstretchfactor}{4}
\providecommand{\BIBentryALTinterwordspacing}{\spaceskip=\fontdimen2\font plus
\BIBentryALTinterwordstretchfactor\fontdimen3\font minus
  \fontdimen4\font\relax}
\providecommand{\BIBforeignlanguage}[2]{{%
\expandafter\ifx\csname l@#1\endcsname\relax
\typeout{** WARNING: IEEEtran.bst: No hyphenation pattern has been}%
\typeout{** loaded for the language `#1'. Using the pattern for}%
\typeout{** the default language instead.}%
\else
\language=\csname l@#1\endcsname
\fi
#2}}
\providecommand{\BIBdecl}{\relax}
\BIBdecl

\bibitem{meiring2015review}
G.~A.~M. Meiring and H.~C. Myburgh, ``A review of intelligent driving style
  analysis systems and related artificial intelligence algorithms,''
  \emph{Sensors}, vol.~15, no.~12, pp. 30\,653--30\,682, 2015.

\bibitem{feng2012selected}
Z.-X. Feng, J.~Liu, Y.-Y. Li, and W.-H. Zhang, ``Selected model and sensitivity
  analysis of aggressive driving behavior,'' \emph{Zhongguo Gonglu Xuebao(China
  Journal of Highway and Transport)}, vol.~25, no.~2, pp. 106--112, 2012.

\bibitem{guy2011simulating}
S.~J. Guy, S.~Kim, M.~C. Lin, and D.~Manocha, ``Simulating heterogeneous crowd
  behaviors using personality trait theory,'' in \emph{Proceedings of the ACM
  SIGGRAPH/Eurographics symposium on computer animation}, 2011, pp. 43--52.

\bibitem{bera2017aggressive}
A.~Bera, T.~Randhavane, and D.~Manocha, ``Aggressive, tense, or shy?
  identifying personality traits from crowd videos,'' in \emph{Proceedings of
  the Twenty-Sixth International Joint Conference on Artificial Intelligence,
  IJCAI-17}, 2017.

\bibitem{Geiger2012CVPR}
A.~Geiger, P.~Lenz, and R.~Urtasun, ``Are we ready for autonomous driving? the
  kitti vision benchmark suite,'' in \emph{Conference on Computer Vision and
  Pattern Recognition (CVPR)}, 2012.

\bibitem{bojarski2016end}
M.~Bojarski, D.~Del~Testa, D.~Dworakowski, B.~Firner, B.~Flepp, P.~Goyal, L.~D.
  Jackel, M.~Monfort, U.~Muller, J.~Zhang \emph{et~al.}, ``End to end learning
  for self-driving cars,'' \emph{arXiv:1604.07316}, 2016.

\bibitem{best2017autonovi}
A.~Best, S.~Narang, L.~Pasqualin, D.~Barber, and D.~Manocha, ``Autonovi:
  Autonomous vehicle planning with dynamic maneuvers and traffic constraints,''
  \emph{arXiv:1703.08561}, 2017.

\bibitem{aljaafreh2012driving}
A.~Aljaafreh, N.~Alshabatat, and M.~S.~N. Al-Din, ``Driving style recognition
  using fuzzy logic,'' in \emph{Vehicular Electronics and Safety (ICVES), 2012
  IEEE International Conference on}, pp. 460--463.

\bibitem{krahe2002predicting}
B.~Krah{\'e} and I.~Fenske, ``Predicting aggressive driving behavior: The role
  of macho personality, age, and power of car,'' \emph{Aggressive Behavior},
  vol.~28, no.~1, pp. 21--29, 2002.

\bibitem{beck2014distress}
K.~H. Beck, B.~Ali, and S.~B. Daughters, ``Distress tolerance as a predictor of
  risky and aggressive driving,'' \emph{Traffic injury prevention}, vol.~15,
  no.~4, pp. 349--354, 2014.

\bibitem{brill2009predictive}
J.~C. Brill, M.~Mouloua, E.~Shirkey, and P.~Alberti, ``Predictive validity of
  the aggressive driver behavior questionnaire (adbq) in a simulated
  environment,'' in \emph{Proceedings of the Human Factors and Ergonomics
  Society Annual Meeting}, vol.~53, no.~18.\hskip 1em plus 0.5em minus
  0.4em\relax SAGE Publications Sage CA: Los Angeles, CA, 2009, pp. 1334--1337.

\bibitem{murphey2009driver}
Y.~L. Murphey, R.~Milton, and L.~Kiliaris, ``Driver's style classification
  using jerk analysis,'' in \emph{Computational Intelligence in Vehicles and
  Vehicular Systems, 2009. CIVVS'09. IEEE Workshop on}, pp. 23--28.

\bibitem{mohamad2011abnormal}
I.~Mohamad, M.~A.~M. Ali, and M.~Ismail, ``Abnormal driving detection using
  real time global positioning system data,'' in \emph{Space Science and
  Communication (IconSpace), 2011 IEEE International Conference on}, pp. 1--6.

\bibitem{qi2015leveraging}
G.~Qi, Y.~Du, J.~Wu, and M.~Xu, ``Leveraging longitudinal driving behaviour
  data with data mining techniques for driving style analysis,'' \emph{IET
  intelligent transport systems}, vol.~9, no.~8, pp. 792--801, 2015.

\bibitem{shi2015evaluating}
B.~Shi, L.~Xu, J.~Hu, Y.~Tang, H.~Jiang, W.~Meng, and H.~Liu, ``Evaluating
  driving styles by normalizing driving behavior based on personalized driver
  modeling,'' \emph{IEEE Transactions on Systems, Man, and Cybernetics:
  Systems}, vol.~45, no.~12, pp. 1502--1508, 2015.

\bibitem{wang2017driving}
W.~Wang, J.~Xi, A.~Chong, and L.~Li, ``Driving style classification using a
  semisupervised support vector machine,'' \emph{IEEE Transactions on
  Human-Machine Systems}, vol.~47, no.~5, pp. 650--660, 2017.

\bibitem{sadigh2014data}
D.~Sadigh, K.~Driggs-Campbell, A.~Puggelli, W.~Li, V.~Shia, R.~Bajcsy, A.~L.
  Sangiovanni-Vincentelli, S.~S. Sastry, and S.~A. Seshia, ``Data-driven
  probabilistic modeling and verification of human driver behavior,'' 2014.

\bibitem{iOnRoad}
\BIBentryALTinterwordspacing
``Ionroad app that makes driving safer.'' [Online]. Available:
  \url{http://www.ionroad.com/}
\BIBentrySTDinterwordspacing

\bibitem{you2013carsafe}
C.-W. You, N.~D. Lane, F.~Chen, R.~Wang, Z.~Chen, T.~J. Bao, M.~Montes-de Oca,
  Y.~Cheng, M.~Lin, L.~Torresani \emph{et~al.}, ``Carsafe app: Alerting drowsy
  and distracted drivers using dual cameras on smartphones,'' in
  \emph{Proceeding of the 11th annual international conference on Mobile
  systems, applications, and services}.\hskip 1em plus 0.5em minus 0.4em\relax
  ACM, 2013, pp. 13--26.

\bibitem{bergasa2014drivesafe}
L.~M. Bergasa, D.~Almer{\'\i}a, J.~Almaz{\'a}n, J.~J. Yebes, and R.~Arroyo,
  ``Drivesafe: An app for alerting inattentive drivers and scoring driving
  behaviors,'' in \emph{Intelligent Vehicles Symposium Proceedings, 2014
  IEEE}.\hskip 1em plus 0.5em minus 0.4em\relax IEEE, 2014, pp. 240--245.

\bibitem{treiber2000congested}
M.~Treiber, A.~Hennecke, and D.~Helbing, ``Congested traffic states in
  empirical observations and microscopic simulations,'' \emph{Physical review
  E}, vol.~62, no.~2, p. 1805, 2000.

\bibitem{kesting2007general}
A.~Kesting, M.~Treiber, and D.~Helbing, ``General lane-changing model mobil for
  car-following models,'' \emph{Transportation Research Record: Journal of the
  Transportation Research Board}, no. 1999, pp. 86--94, 2007.

\bibitem{choudhury2005modeling}
C.~F. Choudhury, ``Modeling lane-changing behavior in presence of exclusive
  lanes,'' Ph.D. dissertation, Massachusetts Institute of Technology, 2005.

\bibitem{katrakazas2015real}
C.~Katrakazas, M.~Quddus, W.-H. Chen, and L.~Deka, ``Real-time motion planning
  methods for autonomous on-road driving: State-of-the-art and future research
  directions,'' \emph{Transportation Research Part C: Emerging Technologies},
  vol.~60, pp. 416--442, 2015.

\bibitem{saifuzzaman2014incorporating}
M.~Saifuzzaman and Z.~Zheng, ``Incorporating human-factors in car-following
  models: a review of recent developments and research needs,''
  \emph{Transportation research part C: emerging technologies}, vol.~48, pp.
  379--403, 2014.

\bibitem{kolski2006autonomous}
S.~Kolski, D.~Ferguson, M.~Bellino, and R.~Siegwart, ``Autonomous driving in
  structured and unstructured environments,'' in \emph{Intelligent Vehicles
  Symposium}.\hskip 1em plus 0.5em minus 0.4em\relax IEEE, 2006, pp. 558--563.

\bibitem{hoffmann2007autonomous}
G.~M. Hoffmann, C.~J. Tomlin, M.~Montemerlo, and S.~Thrun, ``Autonomous
  automobile trajectory tracking for off-road driving: Controller design,
  experimental validation and racing,'' in \emph{American Control Conference
  ACC'07}.\hskip 1em plus 0.5em minus 0.4em\relax IEEE, 2007, pp. 2296--2301.

\bibitem{ziegler2014making}
J.~Ziegler, P.~Bender, M.~Schreiber, H.~Lategahn, T.~Strauss, C.~Stiller,
  T.~Dang, U.~Franke, N.~Appenrodt, C.~G. Keller \emph{et~al.}, ``Making bertha
  drive—an autonomous journey on a historic route,'' \emph{IEEE Intelligent
  Transportation Systems Magazine}, vol.~6, no.~2, pp. 8--20, 2014.

\bibitem{buehler2009darpa}
M.~Buehler, K.~Iagnemma, and S.~Singh, \emph{The DARPA urban challenge:
  autonomous vehicles in city traffic}.\hskip 1em plus 0.5em minus 0.4em\relax
  springer, 2009, vol.~56.

\bibitem{geiger2012team}
A.~Geiger, M.~Lauer, F.~Moosmann, B.~Ranft, H.~Rapp, C.~Stiller, and
  J.~Ziegler, ``Team annieway's entry to the 2011 grand cooperative driving
  challenge,'' \emph{IEEE Transactions on Intelligent Transportation Systems},
  vol.~13, no.~3, pp. 1008--1017, 2012.

\bibitem{kianfar2012design}
R.~Kianfar, B.~Augusto, A.~Ebadighajari, U.~Hakeem, J.~Nilsson, A.~Raza, R.~S.
  Tabar, N.~V. Irukulapati, C.~Englund, P.~Falcone \emph{et~al.}, ``Design and
  experimental validation of a cooperative driving system in the grand
  cooperative driving challenge,'' \emph{IEEE transactions on intelligent
  transportation systems}, vol.~13, no.~3, pp. 994--1007, 2012.

\bibitem{englund2016grand}
C.~Englund, L.~Chen, J.~Ploeg, E.~Semsar-Kazerooni, A.~Voronov, H.~H.
  Bengtsson, and J.~Didoff, ``The grand cooperative driving challenge 2016:
  boosting the introduction of cooperative automated vehicles,'' \emph{IEEE
  Wireless Communications}, vol.~23, no.~4, pp. 146--152, 2016.

\bibitem{sadigh2016planning}
D.~Sadigh, S.~Sastry, S.~A. Seshia, and A.~D. Dragan, ``Planning for autonomous
  cars that leverage effects on human actions.'' in \emph{Robotics: Science and
  Systems}, 2016.

\bibitem{sadigh2016information}
D.~Sadigh, S.~S. Sastry, S.~A. Seshia, and A.~Dragan, ``Information gathering
  actions over human internal state,'' in \emph{Intelligent Robots and Systems
  (IROS), 2016 IEEE/RSJ International Conference on}, pp. 66--73.

\bibitem{huang2017enabling}
S.~H. Huang, D.~Held, P.~Abbeel, and A.~D. Dragan, ``Enabling robots to
  communicate their objectives,'' \emph{arXiv:1702.03465}, 2017.

\bibitem{dorsa2017active}
A.~D.~D. Dorsa~Sadigh, S.~Sastry, and S.~A. Seshia, ``Active preference-based
  learning of reward functions,'' in \emph{Robotics: Science and Systems
  (RSS)}, 2017.

\bibitem{HARRIS20141}
P.~B. Harris, J.~M. Houston, J.~A. Vazquez, J.~A. Smither, A.~Harms, J.~A.
  Dahlke, and D.~A. Sachau, ``The prosocial and aggressive driving inventory
  (padi): A self-report measure of safe and unsafe driving behaviors,''
  \emph{Accident Analysis \& Prevention}, vol.~72, no. Supplement C, pp. 1 --
  8, 2014.

\bibitem{lan2009smartldws}
M.~Lan, M.~Rofouei, S.~Soatto, and M.~Sarrafzadeh, ``Smartldws: A robust and
  scalable lane departure warning system for the smartphones,'' in
  \emph{Intelligent Transportation Systems, 2009. ITSC'09. 12th International
  IEEE Conference on}, pp. 1--6.

\bibitem{i80}
J.~Halkia and J.~Colyar, ``Interstate 80 freeway dataset,'' \emph{Federal
  Highway Administration, U.S. Department of Transportation}, 2006.

\bibitem{bareiss2015generalized}
D.~Bareiss and J.~van~den Berg, ``Generalized reciprocal collision avoidance,''
  \emph{The International Journal of Robotics Research}, vol.~34, no.~12, pp.
  1501--1514, 2015.

\end{thebibliography}

\end{document}